\documentclass{article}%
\usepackage[T1]{fontenc}%
\usepackage[utf8]{inputenc}%
\usepackage{lmodern}%
\usepackage{textcomp}%
\usepackage{lastpage}%
\usepackage{array}
\usepackage{geometry}%
\usepackage{multicol}%
\usepackage{graphicx}%
\usepackage{lipsum}%
\usepackage{caption}%
\usepackage{booktabs}%
\usepackage{ragged2e}%
\usepackage{breqn}  % Pour avoir des équations dmath sur 2 lignes.
\title{Explainable Global Error Weighted on Feature Importance: The \textit{xGEWFI} metric to evaluate the error of data imputation and data augmentation}%
\author{Jean{-}Sébastien Dessureault, Daniel Massicotte}%
\date{\today}%
\begin{document}%
\normalsize%
\maketitle%
\justify%
\section{ABSTRACT}%
\label{sec:ABSTRACT}%
Evaluating the performance of an algorithm is crucial. Evaluating the performance of data imputation and data augmentation can be similar since both generated data can be compared with an original distribution. Although, the typical evaluation metrics have the same flaw: They calculate the feature's error and the global error on the generated data without weighting the error with the feature importance. The result can be good if all of the feature's importance is similar. However, in most cases, the importance of the features is imbalanced, and it can induce an important bias on the features and global errors. This paper proposes a novel metric named "Explainable Global Error Weighted on Feature Importance"(\textit{xGEWFI}). This new metric is tested in a whole preprocessing method that 1. detects the outliers and replaces them with a null value. 2. imputes the data missing, and 3. augments the data. At the end of the process, the \textit{xGEWFI} error is calculated. The distribution error between the original and generated data is calculated using a \textit{Kolmogorov-Smirnov test} (KS test) for each feature. Those results are multiplied by the importance of the respective features, calculated using a \textit{Random Forest} (RF) algorithm. The metric result is expressed in an explainable format, aiming for an ethical AI. \newline

\noindent Keywords: xGEWFI, Data imputation, Data augmentation, Random forest, SMOTE, KNNImputer

\begin{multicols}{2}%
\section{Introduction}%
\label{sec:Introduction}%
% Contexte, description du problème
In a context where missing data imputation and data augmentation are evolving rapidly, it is important to develop new tools to evaluate those machine learning algorithms. This novel method weights the error calculated after generating data (missing or augmented data). The weighting is done on the importance of the features, giving an error a weight related to its feature importance. It is also important to create new tools with the ability to explain the results. For the ethical purpose, from now on, efforts have to be made to produce algorithms with explainable results. This novel \textit{xGEWFI} metric includes itself in this paradigm. 

% Kolmogorov-Smirnov test
To compare the generated data to the original data, the \textit{xGEWFI} metric uses a \textit{Kolmogorov-Smirnov test} \cite{massey_kolmogorov-smirnov_1951}. This test has been widely used for decades in mathematics. It compares the properties of the distribution of two series of data. In a data imputation context, the imputed data for a feature are compared with the original data for this same feature. In a data augmentation context, the added data are compared with the original data. Comparing the generated and the original data, two metrics (statistical and value) are returned to describe the covariance between them. It shows the quality of the data generation process. This test is well-known and has been used since decades \cite{justel_multivariate_1997} \cite{lopes_two-dimensional_2007}
\cite{berger_kolmogorovsmirnov_2014} \cite{fasano_multidimensional_1987}. 

% Random Forest
At the heart of the \textit{xGEWFI} metric is the evaluation of the feature's importance. Normally, all measures of feature error are equally weighted. A better representation would take into account the importance of each feature. This is why a \textit{Random Forest} (RF) algorithm is used in this novel metric. Based on \cite{biau_random_2016} \cite{ronaghan_mathematics_2019} \cite{gulea_how_2019}, an RF algorithm is used as a regressor, as a classifier, but also as a tool to evaluate the importance of the features. It works in a supervised learning context, based on \textit{bayesians networks} and on \textit{decisions trees}. It has been widely used in a variety of applications \cite{lv_simulating_2021}

The next algorithms (\textit{SMOTE}, \textit{KNNImputer}, and \textit{IQR}) are useful to implement the whole method that test the \textit{xGEWFI} algorithm. This method is a whole preprocessing pipeline that makes detection of outliers, data imputation and data augmentation.   

% IQR
The process of outliers detection is made using the \textit{IQR} method. Outliers are feature's values that fall outside of the normal distribution. The \textit{IQR} method defines outliers in a mathematical and formal matter, using the interquartile range rule. It is commonly used as in the papers of \cite{vinutha_detection_2018} \cite{noauthor_interquartile_nodate} \cite{sanchez-gonzalez_median_2020} \cite{wan_estimating_2014}. 

% KNNImputer
The data imputation process is done using the \textit{K-Nearest-Neighbor Imputer} (KNNImputer). It finds a missing value using \textit{k-Nearest Neighbors} algorithm. 
Each sample's missing values are imputed using the mean value from n nearest neighbors availables for a given feature. It is widely used, like in the works of \cite{garcia-laencina_k_2009} \cite{thirukumaran_missing_2012} \cite{tutz_improved_2015} \cite{de_silva_missing_2016}.  There are some more advanced methods to impute some missing data.  Some specialized \textit{Generative Adversial Networks} (GAN), like \textit{Generative Adversial Imputation Networks} (GAIN) algorithm has been recently created  
\cite{noauthor_180602920_nodate} \cite{yoon_gain_2018} \cite{wang_pc-gain_2021}
\cite{popolizio_missing_2021} \cite{zhang_missing_2021}. 

% SMOTE
The data augmentation is implemented using the \textit{Synthetic Minority Oversampling Technique} (SMOTE) algorithm \cite{chawla_smote_2002}. This algorithm help to solve the imbalanced datasets problem by over-sampling the minority classes.  It exists some more recent and state-of-the-art methods to perform data augmentation.  The GANs, for instance \cite{antoniou_data_2017} \cite{Imbalanced} \cite{shorten_survey_2019} uses two deep neural networks to create new data. Although, the SMOTE algorithm is still widely used \cite{hasanin_severely_2019} \cite{rastogi_imbalanced_2018} \cite{demidova_svm_2017}
\cite{guo_improved_2019} and sufficient for this method that validate the proposed \textit{xGEWFI} metric. 
 
% Dataset description
The datasets used to validate this method has been generated using the \textit{Scikit-Learn} framework \cite{kramer_scikit-learn_2016}. Especially the \textit{datasets.make_regression()} \cite{noauthor_sklearndatasetsmake_regression_nodate} and \textit{datasets.make_classification()} \cite{noauthor_sklearndatasetsmake_classification_nodate} functions using different parameters. Both were used to produce a different datasets based on regression and classification.  Both functions are widely used in various paper such as \cite{veugen_privacy-preserving_2021} \cite{civieta_asgl_2021} \cite{ghosh_uncertainty_2021} \cite{guedj_kernel-based_2020}. The generated datasets had the advantage of being fully reproducible when being called with the same parameters.

% Contribution
The main contribution of this paper is to propose a more precise and less biased metric that into account the importance of the feature and, consequently, the importance of the feature's error. It also contributes to AI ethics by proposing an explainable metric, especially for data missing and data augmentation. 

% Structure
The next sections of this paper are organized with the following structure: Section \ref{sec:Methodology} describes the proposed methodology. Section \ref{sec:Results} presents the results. Section \ref{sec:Discussions} discusses about the results and their meaning and Section \ref{sec:Conclusion} concludes this research.

\section{Methodology}%
\label{sec:Methodology}%

\subsection{Preprocessing of dataset}%
\label{subsec:Sub_methodology_preprocess}%
% Description plus complète du dataset.
As mention in \ref{sec:Introduction}, the datasets are generated by  
the \textit{datasets.make_regression()} and the \textit{datasets.make_classification()} of the\textit{ Scikit-learn} framework.  When called, the parameters allow selecting different dataset characteristics according to the test. It is possible to customize the number of rows and features. Here is a description of the parameters that generate the synthetic data. \textit{n_sample}: The number of rows or generated data.  \textit{n_features}: The number of columns or features. \textit{n_target} and \textit{n_classes}: The number of field that targets the regression and the classification data respectively. In all our cases, this one is always equal to 1. \textit{shuffle}: When equal to True, it randomly changes the order of the data. This parameter is always True in our cases.  \textit{random_state}: It is the seed used to randomly generated the data. If the same seed (an integer number) is used, then the generated data will always be the same. In our case, a value of 1 is always used for better reproducibility. 
For this study, 5\% of outliers and 30\% of data missing have been randomly generated.

\subsection{Architecture}%
\label{subsec:Sub_methodology_architecture}%

% Description de l'architecture
Figure \ref{fig:Archi} shows the preprocessing method's architecture that includes and validates the \textit{xGEWFI} metric.
\noindent\begin{center} \includegraphics[width=\columnwidth]{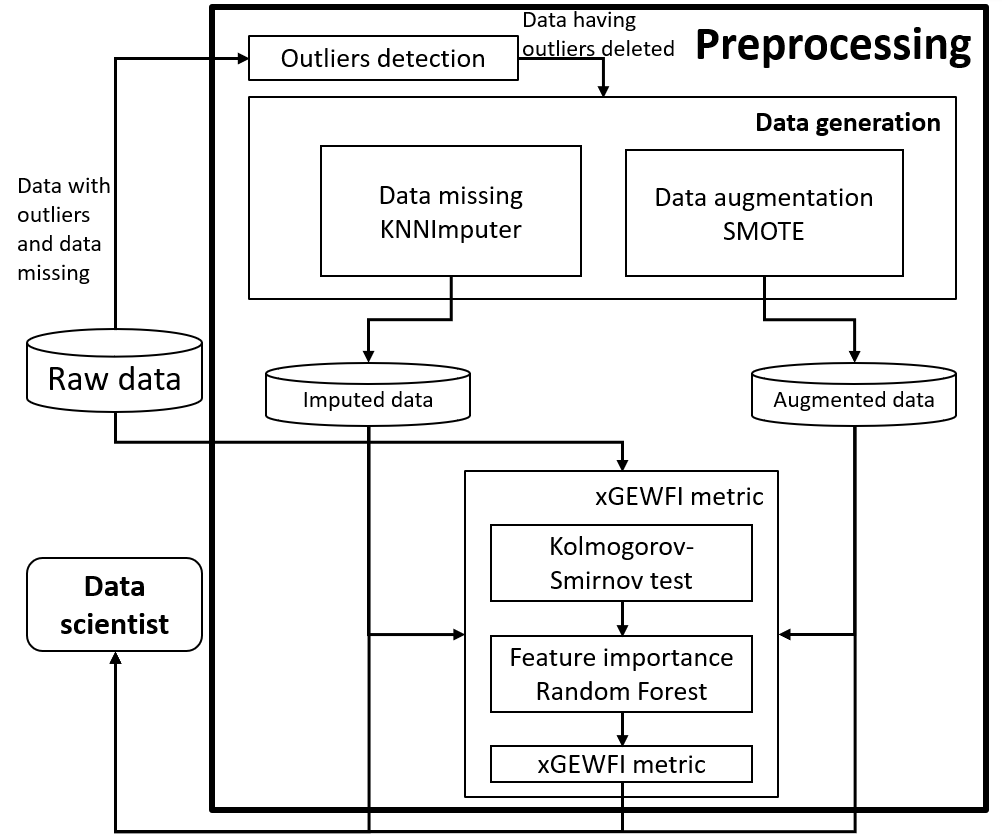}%
\captionof{figure}{Architecture of the preprocessing  method that includes and validates the \textit{xGEWFI} metric \label{fig:Archi}}
\end{center}%

This figure shows that the first step consists of sending the raw data to the outliers detection module. This part does the following 1. detects outliers 2. replaces the outliers with a null value. Those null values will be replaced by correct values at the next step, using a KNNImputer Algorithm. 
The next part does the data generation. The first one imputes the data missing using a KNNImputer algorithm. The second one does the data augmentation process using a SMOTE algorithm. The results are the imputed data and the augmented data generated. Both are returned to the data scientist and sent to the \textit{xGEWFI} algorithm. The \textit{xGEWFI} algorithm compares the generated data distribution to the original data's distribution using a KS test. It evaluates the error of the data generation process, both globally and for each feature. An RF algorithm is executed afterward to find the feature importances. Finally, the \textit{xGEWFI} is calculated based on each feature's error and importance. All the information for the \textit{xGEWFI} metric (error and explainability) is sent to the data scientist.

\subsection{\textit{xGEWFI} metric - Global Error Weighted on Feature Importance}%
\label{subsec:Sub_methodology_metric}%
% https://towardsdatascience.com/the-mathematics-of-decision-trees-random-forest-and-feature-importance-in-scikit-learn-and-spark-f2861df67e3
% https://machinelearningmastery.com/calculate-feature-importance-with-python/

Metrics usually measure errors or performances of a process without considering the feature's importance. The premise of \textit{xGEWFI} metric is that the feature importance should weigh the error amplitude. A high-value error should represent nothing if the feature importance is null or extremely low. Conversely, the amplitude of an error should be higher with high feature importance. 
We must first calculate the feature importance using an RF algorithm to calculate the \textit{xGEWFI} metric. RF algorithms are based on Bayesian networks and multiple decision trees. A single decision tree may lead to some bias, so having tenths of decision trees (as proposed by the RF algorithm) solve this problem, as it pools all the outcomes to return the most frequent answer. It exists two applications of RF algorithms: 1. RF regressors and 2. classifiers. Both are also used to evaluate the importance of the features. The solution can be displayed in a tree graph where the nodes represent the decision. The branches represent the possible outcomes. The leaves represent all the possible answers, combined with their probability. With this ability to compute a tree of
bayesian probability, the RF algorithms can also calculate the feature importances in the regression of the classification process. The importance of each feature is given in a normalized form. 

%https://www.tutorialspoint.com/statistics/kolmogorov_smirnov_test.htm#:~:text=Fo(X)%20%3D%20Observed,of%20observations).
Then we must calculate the performance of the data generation algorithms (data imputation and data augmentation). The KS test is used to do so. 
This test is used to compare two distributions. In our case, the original data distribution is compared with the generated (imputed or augmented) data distribution. The formula of the KS test is shown in \ref{eq:KS1}. \newline

\begin{dmath}
D_{f} = sup | F_{o,f}(x) - F_{g,f}(x) | 
\label{eq:KS1}
\end{dmath}%

Where $D_{f}$ is the result, the D-statistic, for feature \textit{f}.  $F_{o,f}$ is the distribution function of the feature \textit{f} of \textit{o}, the original \textit{x} data.  Similarly, $F_{g,f}$ is the distribution function of the same feature \textit{f} of \textit{g}, the generated \textit{x} data.  The case where there is no difference between the two distributions is called the null hypothesis.  In \ref{eq:KS1} the result D (the D-statistic) is the evaluation of the error.  A value of $D = 0.0$ is the null hypothesis. 

Without calculating the feature importance, we can conclude that, since the D-statistics indicates the error between two distributions, it can be considered as the error of the generated distribution as in \ref{eq:FeatureError}.

\begin{dmath}
E_{f} = D_{f} 
\label{eq:FeatureError}
\end{dmath}%

The global error is the sum of all the feature's errors, as in \ref{eq:GlobalError}

\begin{dmath}
Global Error_{f} = \sum_{f=1}^{featureNb.}{E_{f}} 
\label{eq:GlobalError}
\end{dmath}%

Even though those numbers are not the finality, they are returned by the \textit{xGEWFI} algorithm as part of the final answer for explainability matters. Those numbers are an important part of the final answer and must be available to the data scientist to help him to have a better comprehension of the process.  

Based on both the RF algorithm and the KS test, the \textit{xGEWFI} metric relies on the D-Statistics and on the importance of the features. \ref{eq:FeatureWeightedError} and \ref{eq:GlobalWeightedError} shows the calculations for each feature error and for global error, respectively. 

\begin{dmath}
Weighted Error_{f} = E_{f} * W_{f} 
\label{eq:FeatureWeightedError}
\end{dmath}%

\textit{f} is the feature index. $E_{f}$ is the error of feature \textit{f} using the KS test. $W_{f}$ is the weight of feature \textit{f} according to the RF algorithm. \newline

\begin{dmath}
xGEWFI = \sum_{f=1}^{featureNb.}{Weighted Error_{f}}
\label{eq:GlobalWeightedError}
\end{dmath}%

\subsection{Outliers data}%
\label{subsec:Sub_methodology_outliers}%

%Inter quartile range (IQR) method
%https://towardsdatascience.com/practical-implementation-of-outlier-detection-in-python-90680453b3ce

The first part of the process consists of identifying the data outliers. The need is to identify the outliers and replace them with some null values that will be processed later. To do that part, we use the IQR method. It consists of dividing each feature of the dataset into quartiles. Fig. \ref{fig:IQR} shows a box plot representation of the IQR method.\newline

\noindent \includegraphics[width=\columnwidth]{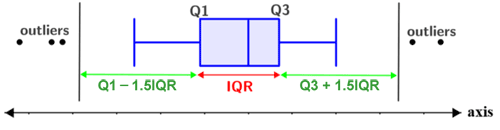}%
\captionof{figure}{IQR method to detect outliers \cite{noauthor_interquartile_nodate}\newline
 \label{fig:IQR}}%
 
The first quartile point Q1 indicates that 25\% of the data points are below that value. The second quartile Q2 is the median point of the feature. The third quartile Q3 is at the point where 75\% of the data points are below that value. Eq. \ref{eq:IQR} defines the IQR, \ref{eq:LLimit} defines the lower limit, and \ref{eq:ULimit} defines the upper limit. 

\begin{dmath}
IQR = Q3 - Q1
\label{eq:IQR}
\end{dmath}%

\begin{dmath}
Lower Limit = Q1 - 1.5 * IQR
\label{eq:LLimit}
\end{dmath}%

\begin{dmath}
Upper Limit = Q3 + 1.5 * IQR
\label{eq:ULimit}
\end{dmath}%

When a value outside the lower and upper is found, it is replaced by a null value. This value will be processed once again in the next step while processing the missing data. 

\subsection{Data imputation}%
\label{subsec:Sub_methodology_missing}%
The system trained a KNNImputer to impute the values of the missing data. This algorithm relies on the KNN algorithm aiming to find the k nearest neighbour of the missing data. This algorithm is based on the computation of the distance between the data. In this case, the euclidian distance is used as in \ref{eq:Euclidian}.

%https://www.saedsayad.com/k_nearest_neighbors.htm
\begin{dmath}
D = \sqrt{\sum_{i=1}^{k}{\left( x_{i} - y_{i} \right)^{2}}}
\label{eq:Euclidian}
\end{dmath}%

\textit{D} is the euclidian distance, \textit{k} is the number of neighbours to find.  Finally, \textit{x} and \textit{y} are the origin and destination data, respectively. 

\subsection{Data augmentation}%
\label{subsec:Sub_methodology_augmentation}%
%https://www.geeksforgeeks.org/ml-handling-imbalanced-data-with-smote-and-near-miss-algorithm-in-python/
A SMOTE architecture is used to oversample data of the original dataset. It helps to solve the imbalanced data problem. It balances class distribution by randomly increasing minority instances. Virtual training records are generated by linear interpolation for the minority class. The algorithm selects the k-nearest neighbours for each data added in the minority class. To compute the distance between the neighbours, the euclidian distance is used as defined in \ref{eq:Euclidian}.

\section{Results}%
\label{sec:Results}%

Let us study two cases to explain the advantages of the \textit{xGEWFI} metric that weights the error with the feature's importance. In a case where all the features have the same importance, the weighting specific to the \textit{xGEWFI} process would not affect the result much. Although, in the majority of the cases, the difference is significant. The following subsections present two typical and common cases where the \textit{xGEWFI} metric changes drastically our evaluation of the data imputation and data augmentation process. Both cases use a different dataset. Both datasets have five features and 25000 data.  \newline

\subsection{Case 1 - Similar features errors and different features importances}%
\label{subsec:Case1}%

% Description of the problem
This first case evaluates the results of a data imputation on a regression problem. We have a similar feature error as presented in Fig. \ref{fig:STD_case1}. We can see in this figure that every feature shows an error of near 0.35 on the KS test (axe Y) for each feature on axe X.

\vfill\null
\columnbreak

% KS STD
\noindent \includegraphics[width=\columnwidth]{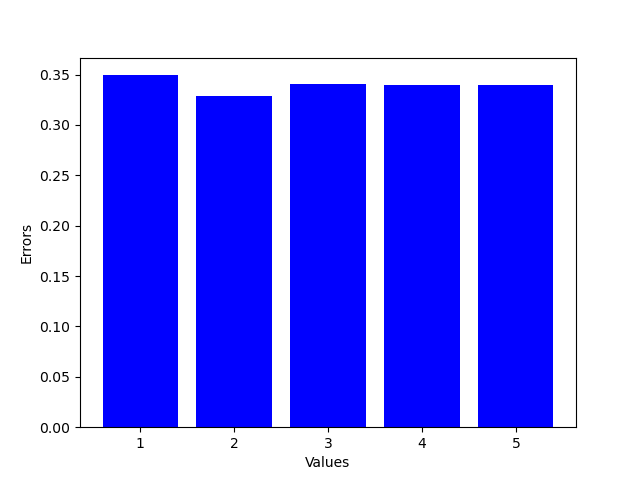}%
\captionof{figure}{Standard KS test error.\newline \label{fig:STD_case1}}

The sum of the errors is 1.68, representing a final metric on the quality of the imputation process. Without the xGEWFI metric, evaluating the data imputation would stop here. Is this metric alone the best method to represent the quality of this data imputation process? The \textit{xGEWFI} method proposes a method that considers each feature's weight.   The final result (the \textit{xGEWFI} error) will be more representative of the quality of the augmentation/imputation process, having weighted the KS test error on the feature importance. For this case, Fig. \ref{fig:importance_case1} presents the importance of the features. We can see that every feature on axe X has a different level of normalized importance on axe Y. 

% Importance
\noindent \includegraphics[width=\columnwidth]{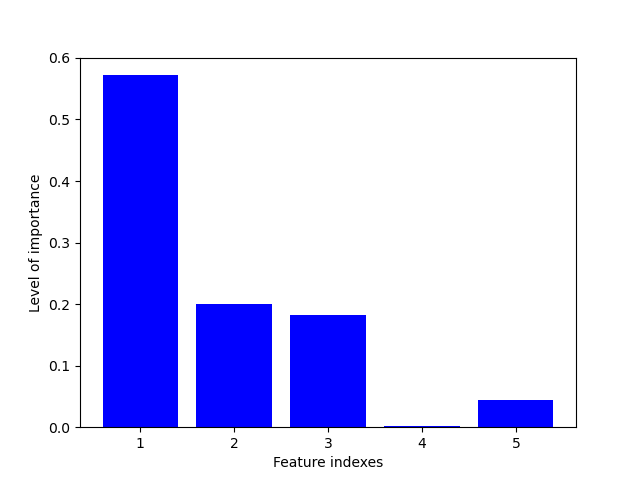}%
\captionof{figure}{Importance of the features.\newline \label{fig:importance_case1}}%

Fig. \ref{fig:xGEWFI_case1} shows the result of the KS test weighted by the feature importance.  Since every feature has about the same error as shown in Fig. \ref{fig:STD_case1}, the result of the \textit{xGEWFI} error is also about the same as the feature importance.  It can be seen clearly by comparing Fig. \ref{fig:importance_case1} and Fig. \ref{fig:xGEWFI_case1}.

% xGEWFI
\noindent \includegraphics[width=\columnwidth]{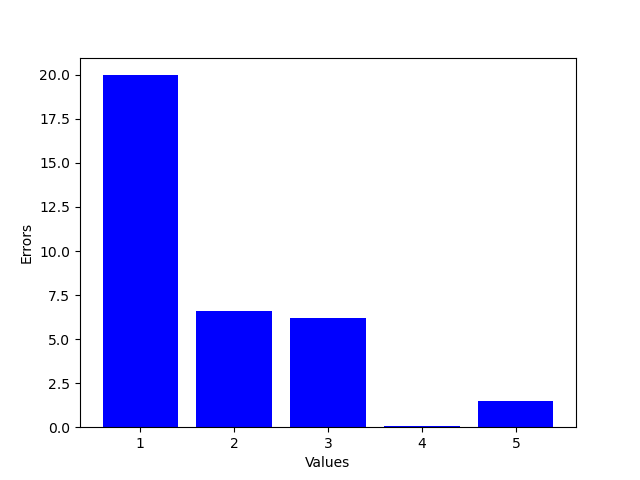}%
\captionof{figure}{Weighted \textit{xGEWFI} error.\newline \label{fig:xGEWFI_case1}}%

% Conclusion
We can conclude for this test that the error using a KS test is very different than the \textit{xGEWFI} error (Fig. \ref{fig:STD_case1} and Fig. \ref{fig:xGEWFI_case1}). The global error of the KS test was 1.68, and the \textit{xGEWFI} global error was 33.75. The magnitude of the two metrics cannot be compared. Each one must be compared with other errors of the same kind. 

\subsection{Case 2 - Important differences between features importances and errors}%
\label{subsec:Case2}%

% Description of the problem
This second case evaluates the results of data augmentation on a classification problem.  Fig. \ref{fig:STD_case2} shows the result of the KS test representing the error (ax Y) on each feature (ax X).  This figure shows an important variation between each feature's error at the opposite of the KS test error presented in case 1.     

% KS STD
\noindent \includegraphics[width=\columnwidth]{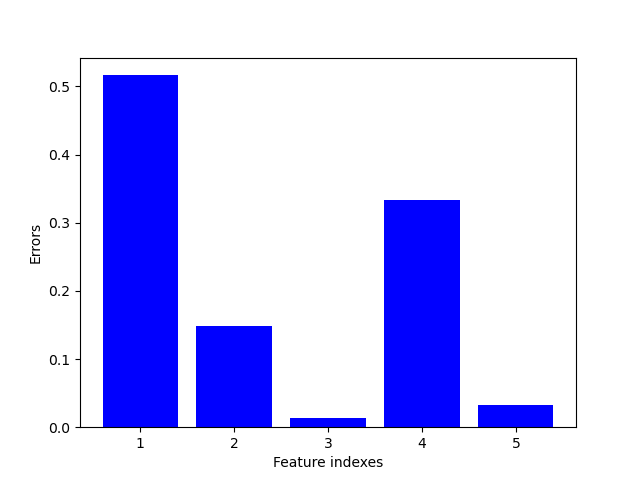}%
\captionof{figure}{Standard KS test error.\newline \label{fig:STD_case2}}%

The feature's importance varies significantly, as shown in Fig. \ref{fig:importance_case2}. Without going further using \textit{xGEWFI}, we would conclude that the errors on features 1 and 4 are higher. The errors on features 3 and 5 are insignificant compared to the other feature's errors. A better way to evaluate the error would be to consider the features' importance. Fig. \ref{fig:importance_case2} presents the importance of the features for case 2. 

% Importance
\noindent \includegraphics[width=\columnwidth]{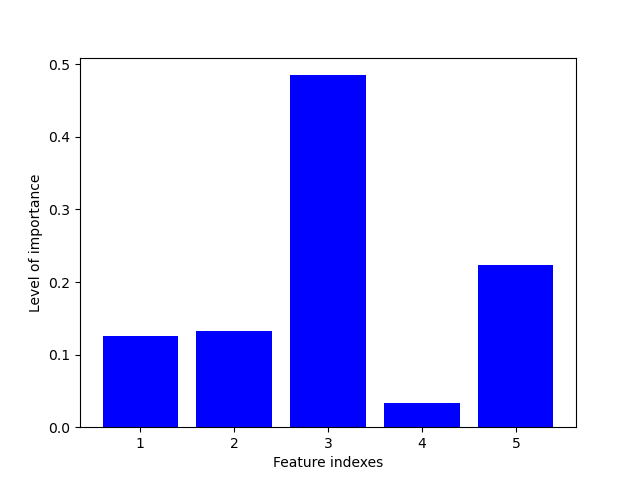}%
\captionof{figure}{Importance of the features.\newline \label{fig:importance_case2}}%

In this figure, the X axe is the dataset's features, and the Y axe is the normalized level of importance of the features.  We can conclude, for instance, that feature 3  is by far the most important, and feature 4 is negligible.  Let us see the impact of weighting the KS test error with the feature's importance.  This is shown in the \textit{xGEWFI} graphic in Fig.\ref{fig:xGEWFI_case2}.

%  xGEWFI
\noindent \includegraphics[width=\columnwidth]{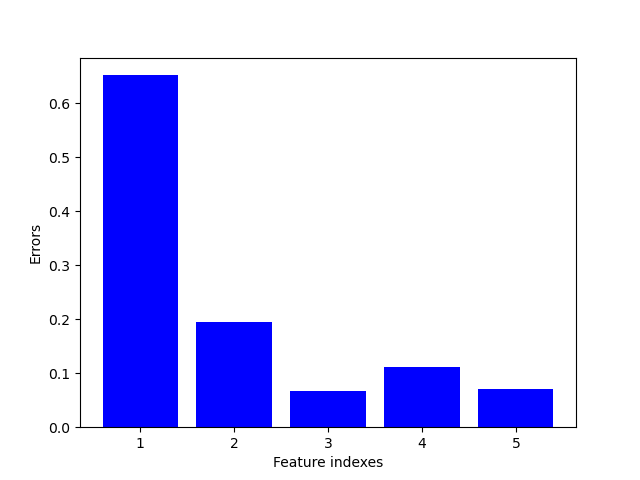}%
\captionof{figure}{Weighted \textit{xGEWFI} error.\newline \label{fig:xGEWFI_case2}}%

We can note that every feature's error has been weighted according to the importance of the features. Let us study specifically feature 3 and feature 4. Feature 3 KS test error (Fig.\ref{fig:STD_case2}) is very low. It is also by far the most important feature (Fig. \ref{fig:importance_case2}. Ultimately, the relative importance of the feature compared to the other features has been raised. As for feature 4, the KS test error was quite high (Fig.\ref{fig:STD_case2}). Since the importance of this feature is very low (Fig. \ref{fig:importance_case2}), the \textit{xGEWFI} error for this feature is lower than its KS test error evaluation. 

% Conclusion
Like in case 1, the second case allows us to conclude that the results are different, whether they are weighted or not using the feature importance using the \textit{xGEWFI} error (Fig. \ref{fig:STD_case1} and Fig. \ref{fig:xGEWFI_case1}). 
In this case, the global error of the KS test was 0.60, and the \textit{xGEWFI} global error was 0.54. As mentioned in \ref{subsec:Case1}, the magnitude of the two metrics cannot be compared. They must be compared with their respective kind.

This case showed that \textit{xGEWFI} metric as the advantage of being weighted using the feature's importance. Also, the results are explainable, as shown in Section \ref{subsec:SubExplainability}.

\subsection{Explainability}%
\label{subsec:SubExplainability}%

The "x" at the beginning of \textit{xGEWFI} stands for "explainable". This metric has not only been developed to give a quantitative evaluation of a data imputation or a data augmentation. It has also been developed to explain its process. That is why graphics and tables were encapsulated in the \textit{xGEWFI} method. Fig. \ref{fig:STD_case2} \ref{fig:importance_case2}), \ref{fig:xGEWFI_case2}, for instance, are fully part of the method, aiming to give the users a better comprehension of the process. There are also 3 other types of graphics created to help the user better comprehend the data and the process. Fig. \ref{fig:boxplot_case1} is an example (the dataset used for case 1) of a graphic that shows the distribution of each feature using a box plot graphic. 

% Distribution
\noindent \includegraphics[width=\columnwidth]{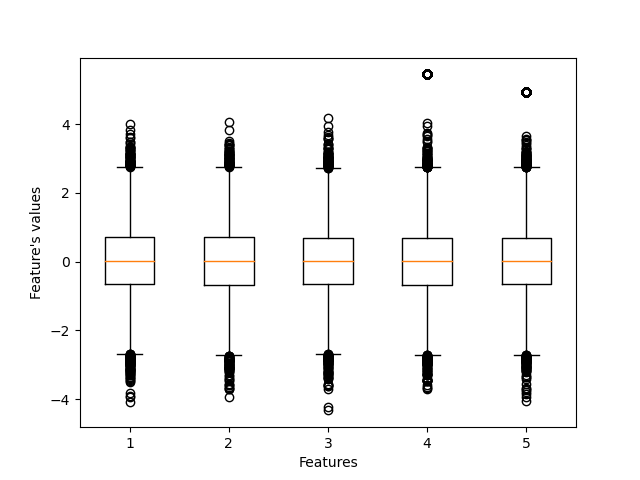}%
\captionof{figure}{Example of a box plot graphic encapsulated in the \textit{xGEWFI} method for better explainability.\newline \label{fig:boxplot_case1}}%

The median marks the mid-point of the feature's values and is shown by the line that divides the box into two parts. Half the scores are greater than or equal to this value, and half are lower. The middle box represents the middle 50\% of scores for the feature. The upper and lower whiskers represent scores outside the middle 50\%. Dots outside of the whiskers are outliers.  

Fig.\ref{fig:distributionKolmogorov_case1} is another visual tool used to explain the data and the process. The \textit{xGEWFI} method generates one graphic of this type by feature. It is a histogram of the original and generated (imputed or augmented) data.  

% KS
\noindent \includegraphics[width=\columnwidth]{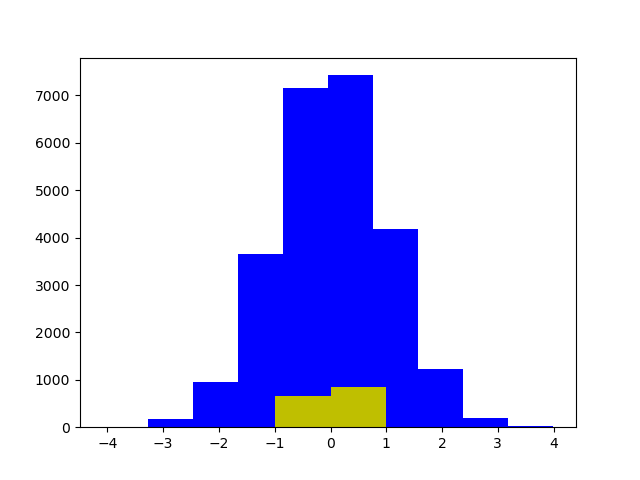}%
\captionof{figure}{Distribution of the feature 1 of the dataset used in case 1.  \newline \label{fig:distributionKolmogorov_case1}}%
Blue bars represent the original data, and yellow bars represent the generated data. X axe is the standard deviation of the feature, and Y axe is the number of occurrences of those standard deviation ranges. Gaussian distribution is visible for both generated and original data. 

% Combo
The third type of graphic combines the 3 important parameters used in the \textit{xGEWFI} equation (\eqref{eq:FeatureWeightedError}. It shows the magnitude of the 3 parameters. 

\vfill\null
\columnbreak

\noindent \includegraphics[width=\columnwidth]{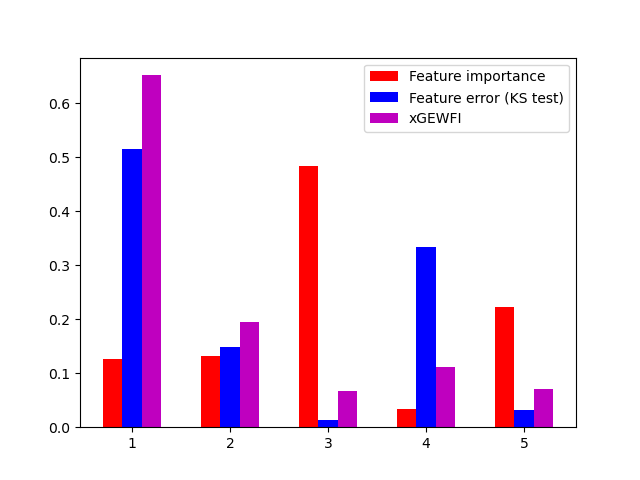}%
\captionof{figure}{Presentation of the 3 important parameters in the \textit{xGEWFI} formula.\newline \label{fig:combo_case1}}%

X axe shows the features, and Y axe shows the magnitude of the values. Red bars are the feature importances, blue bars are the KS test feature errors, and the magenta bar (combines red and blue) are the resulting \textit{xGEWFI} error.\newline 

Ultimately, the method offers the user the numerical data required to calculate the \textit{xGEWFI} metric. The data are displayed on the screen, and two LaTeX tables are generated and ready to be included in a document. For instance, tables \ref{TblxGEWFI1} and \ref{TblxGEWFI2} has been generated by the \textit{xGEWFI} method in case 1. \newline

\begin{minipage}[htb]{1.0\columnwidth}
\caption{Results of the \textit{xGEWFI} metric.}%
\label{TblxGEWFI1}%
\begin{tabular}{l|r}%
Metrics&Values\\%
\hline%
\textit{xGEWFI} mean error&33.76\\%
KS mean error&1.68\\%
\end{tabular}%
\end{minipage}\newline\newline

\begin{minipage}[htb]{1.0\columnwidth}
\caption{Explanability of the \textit{xGEWFI} metric.}%
\label{TblxGEWFI2}%
\begin{tabular}{l|r|r|r}%
Features&Imp.&KS error&\textit{xGEWFI} error\\%
\hline%
Feature 1&0.57&0.34&19.23\\%
Feature 2&0.2&0.34&6.84\\%
Feature 3&0.18&0.34&6.16\\%
Feature 4&0.0&0.34&0.07\\%
Feature 5&0.04&0.33&1.45\\%
\end{tabular}%
\end{minipage}\newline

\subsection{Comparison between this novel method and original methods}%
\label{subsec:Subcomparison}%
KS test has been widely used for decades in statistics. It aims to determine the degree to which two samples follow the same distribution. This method is a simple way to compare the original data to the generated (imputed or augmented) data. This evaluation method has been used in this work since it is known as a reference in statistics to compare two data distributions. This proposed \textit{xGEWFI} method principle could also apply to any other metric. 

Some quantified comparisons between the KS test and the \textit{xGEWFI} method are shown in last section, especially in Fig. \ref{fig:combo_case1}, Table \ref{TblxGEWFI1}, and \ref{TblxGEWFI2}. 

The results show that the evaluation method used (the KS test) will give an evaluation of the similarity between the original data and the generated data for each feature. The \textit{xGEWFI} method weights the results of the KS test using the importance of each feature given by an RF algorithm. We have better feature errors and global error evaluations when those are adjusted according to their importance. In other words, a feature error should be minimized when this feature is less valuable to differentiate the data. Conversely, a feature error should be maximized when this feature is handy to differentiate the data. Having highly differentiable data makes the better classification and regression processes. 

The values returned by this \textit{xGEWFI} metric cannot be directly compared with the other metrics like the KS test. They do not have the same magnitude. It cannot be seen as a disadvantage since most metric magnitudes (like the KS test) are neither compatible with others.  

Finally, a traditional method like the KS test is explainable since it is a statistical formula. Although, when we use it combined with a machine learning algorithm like RF, the results are less explainable. We must use some strategies to explain why this metric gives those results. The \textit{xGEWFI} method is designed for explainability.

\section{Discussions}%
\label{sec:Discussions}%
A novel explainable metric for data imputation and data augmentation has been developed in this paper. To validate this metric, a complete method to detect and correct outliers, impute data and augment data has been made. It has been tested on two datasets of 25000 rows and 5 features. One dataset is a classification problem, and the other is a regression problem. Both datasets are fully reproducible using the \textit{make_regression()} and \textit{make_classification()} functions available in the \textit{scikit-learn} framework. 

The two significant advantages of the \textit{xGEWFI} metric are 1. The feature importance weights each feature error (composing the global error). 2. The result is explainable, as mentioned in section \ref{subsec:SubExplainability}. 

The results are presented in two different cases. The first case was a data imputation problem for a regression-oriented dataset. The KS test gives the same error to all the features (Fig. \ref{fig:STD_case1}). The importance of the features is unequally distributed, as shown in Fig. \ref{fig:importance_case1}. It is obvious that feature 1 is very significant, and feature 4 is insignificant. Since the process gives similar KS test errors for all features, the results of the \textit{xGEWFI} algorithm (\ref{fig:xGEWFI_case1}) will be very similar to the feature importance (Fig. \ref{fig:STD_case1}).  Although, the \textit{xGEWFI} made a huge improvement, since it is the Fig. \ref{fig:STD_case1} and Fig. \ref{fig:xGEWFI_case1} that must be compared. 

The second case was a problem of data augmentation in a classification-oriented dataset. The KS test errors differed for all features (Fig. \ref{fig:STD_case2}). The importance of the features was also different, as presented in Fig. \ref{fig:importance_case2}. The results are shown in Fig. \ref{fig:xGEWFI_case2}.  It represents the KS test values weighted by the feature importances.

There is only one situation where the \textit{xGEWFI} method would not significantly improve the metric. That is when the feature importances are equally distributed between them. Nonetheless, this situation is the exception and not the rule. Hence, the improvement of the \textit{xGEWFI} method is significant in most situations. 

This novel method helps evaluate the performance of a data imputation or augmentation process. A data scientist could try different meta-parameters on a data imputation algorithm (like KNNImpute or GAIN) or a data augmentation algorithm (like SMOTE or GAN). The \textit{xGEWFI} metric will give him an explainable evaluation, weighted on the importance of the feature specific to his dataset.  

In the end, this research succeeds in improving data imputation and data augmentation metrics developed in an explainable manner.

\section{Conclusion}%
\label{sec:Conclusion}%

This paper proposed a novel explainable metric to evaluate the performance of any data imputation and data augmentation method. Using a classification-oriented dataset, or a regression-oriented dataset, this research implements a whole process to 1. Detect outliers and replace them with null values. 2. Impute missing data, and 3. Augment the data. At the end of this process, the proposed \textit{xGEWFI} algorithm computes the error based on the KS test, a well-known statistical reference aiming to compare two feature's distribution (the original one and the generated one). The results of this KS Test are then multiplied by the importance of the respective features generated by an RF algorithm, resulting in a weighted error more representative than the non-weighted error. This result, the \textit{xGEWFI} metric, allows the data scientist to evaluate better the impact of the meta-parameters used in the imputation and augmentation process and on his datasets. 

This method can be improved in the future by replacing the classic KS test with another metric. KS test is a well-known classic method to compare the distribution of two variables. In this context, it has been utilized to test de validity of the imputation and the augmentation process. A more practical test may be found for \textit{xGEWFI} in the future. 
In machine learning, the concept of explainability is relatively recent. This new field is proliferating and will keep growing in the following years. New concepts and methods will flourish, and it is very likely that \textit{xGEWFI} explainability could be improved shortly.

\section{Acknowledment}%
\label{sec:Acknowledment}%
This work has been supported by the "Cellule d’expertise en robotique et intelligence artificielle" of the Cégep de Trois{-}Rivières.

%
%\section{References}%
%\label{sec:References}%
\bibliographystyle{plain}%
\bibliography{paper}

\begin{thebibliography}{10}

\bibitem{noauthor_180602920_nodate}
[1806.02920] {GAIN}: Missing data imputation using generative adversarial nets.

\bibitem{noauthor_interquartile_nodate}
The interquartile range: Theory and estimation - {ProQuest}.

\bibitem{noauthor_sklearndatasetsmake_classification_nodate}
sklearn.datasets.make\_classification.

\bibitem{noauthor_sklearndatasetsmake_regression_nodate}
sklearn.datasets.make\_regression.

\bibitem{antoniou_data_2017}
Antreas Antoniou, Amos Storkey, and Harrison Edwards.
\newblock Data augmentation generative adversarial networks.

\bibitem{berger_kolmogorovsmirnov_2014}
Vance~W. Berger and {YanYan} Zhou.
\newblock Kolmogorov–smirnov test: Overview.
\newblock In {\em Wiley {StatsRef}: Statistics Reference Online}. John Wiley \&
  Sons, Ltd.

\bibitem{biau_random_2016}
Gérard Biau and Erwan Scornet.
\newblock A random forest guided tour.
\newblock 25(2):197--227.
\newblock Company: Springer Distributor: Springer Institution: Springer Label:
  Springer Number: 2 Publisher: Springer Berlin Heidelberg.

\bibitem{chawla_smote_2002}
Nitesh~V. Chawla, Kevin~W. Bowyer, Lawrence~O. Hall, and W.~Philip Kegelmeyer.
\newblock {SMOTE}: synthetic minority over-sampling technique.
\newblock 16(1):321--357.

\bibitem{civieta_asgl_2021}
Álvaro~Méndez Civieta, M.~Carmen Aguilera-Morillo, and Rosa~E. Lillo.
\newblock Asgl: A python package for penalized linear and quantile regression.

\bibitem{de_silva_missing_2016}
Hiroshi de~Silva and A.~Shehan Perera.
\newblock Missing data imputation using evolutionary k- nearest neighbor
  algorithm for gene expression data.
\newblock In {\em 2016 Sixteenth International Conference on Advances in {ICT}
  for Emerging Regions ({ICTer})}, pages 141--146.
\newblock {ISSN}: 2472-7598.

\bibitem{demidova_svm_2017}
Liliya Demidova and Irina Klyueva.
\newblock {SVM} classification: Optimization with the {SMOTE} algorithm for the
  class imbalance problem.
\newblock In {\em 2017 6th Mediterranean Conference on Embedded Computing
  ({MECO})}, pages 1--4.

\bibitem{fasano_multidimensional_1987}
G.~Fasano and A.~Franceschini.
\newblock A multidimensional version of the kolmogorov–smirnov test.
\newblock 225(1):155--170.

\bibitem{garcia-laencina_k_2009}
Pedro~J. García-Laencina, José-Luis Sancho-Gómez, Aníbal~R.
  Figueiras-Vidal, and Michel Verleysen.
\newblock K nearest neighbours with mutual information for simultaneous
  classification and missing data imputation.
\newblock 72(7):1483--1493.

\bibitem{ghosh_uncertainty_2021}
Soumya Ghosh, Q.~Vera Liao, Karthikeyan~Natesan Ramamurthy, Jiri Navratil,
  Prasanna Sattigeri, Kush~R. Varshney, and Yunfeng Zhang.
\newblock Uncertainty quantification 360: A holistic toolkit for quantifying
  and communicating the uncertainty of {AI}.

\bibitem{guedj_kernel-based_2020}
Benjamin Guedj and Bhargav Srinivasa~Desikan.
\newblock Kernel-based ensemble learning in python.
\newblock 11(2):63.
\newblock Number: 2 Publisher: Multidisciplinary Digital Publishing Institute.

\bibitem{gulea_how_2019}
Toma Gulea.
\newblock How not to use random forest.

\bibitem{guo_improved_2019}
Shikai Guo, Yaqing Liu, Rong Chen, Xiao Sun, and Xiangxin Wang.
\newblock Improved {SMOTE} algorithm to deal with imbalanced activity classes
  in smart homes.
\newblock 50(2):1503--1526.

\bibitem{Imbalanced}
Baokun Han, Sixiang Jia, Guifang Liu, and Jinrui Wang.
\newblock Imbalanced fault classification of bearing via wasserstein generative
  adversarial networks with gradient penalty.
\newblock {\em Shock and Vibration}, pages 1 -- 14, 2020.

\bibitem{hasanin_severely_2019}
Tawfiq Hasanin, Taghi~M. Khoshgoftaar, Joffrey~L. Leevy, and Richard~A. Bauder.
\newblock Severely imbalanced big data challenges: investigating data sampling
  approaches.
\newblock 6(1):107.

\bibitem{justel_multivariate_1997}
Ana Justel, Daniel Peña, and Rubén Zamar.
\newblock A multivariate kolmogorov-smirnov test of goodness of fit.
\newblock 35(3):251--259.

\bibitem{kramer_scikit-learn_2016}
Oliver Kramer.
\newblock Scikit-learn.
\newblock pages 45--53.

\bibitem{lopes_two-dimensional_2007}
R.~H.~C. Lopes, I.~D. Reid, and P.~R. Hobson.
\newblock The two-dimensional kolmogorov-smirnov test.
\newblock Proceedings of Science.
\newblock Accepted: 2007-08-15T16:01:55Z.

\bibitem{lv_simulating_2021}
Jianjun Lv, Yifan Wang, Xun Liang, Yao Yao, Teng Ma, and Qingfeng Guan.
\newblock Simulating urban expansion by incorporating an integrated
  gravitational field model into a demand-driven random forest-cellular
  automata model.
\newblock 109:103044.

\bibitem{massey_kolmogorov-smirnov_1951}
Frank~J. Massey.
\newblock The kolmogorov-smirnov test for goodness of fit.
\newblock 46(253):68--78.
\newblock Publisher: Taylor \& Francis.

\bibitem{popolizio_missing_2021}
Marina Popolizio, Alberto Amato, Tiziano Politi, Roberto Calienno, and Vincenzo
  Di~Lecce.
\newblock Missing data imputation in meteorological datasets with the {GAIN}
  method.
\newblock In {\em 2021 {IEEE} International Workshop on Metrology for Industry
  4.0 {IoT} ({MetroInd}4.0 {IoT})}, pages 556--560.

\bibitem{rastogi_imbalanced_2018}
Avnish~Kumar Rastogi, Nitin Narang, and Zamir~Ahmad Siddiqui.
\newblock Imbalanced big data classification: a distributed implementation of
  {SMOTE}.
\newblock In {\em Proceedings of the Workshop Program of the 19th International
  Conference on Distributed Computing and Networking}, Workshops {ICDCN} '18,
  pages 1--6. Association for Computing Machinery.

\bibitem{ronaghan_mathematics_2019}
Stacey Ronaghan.
\newblock The mathematics of decision trees, random forest and feature
  importance in scikit-learn and spark.

\bibitem{shorten_survey_2019}
Connor Shorten and Taghi~M. Khoshgoftaar.
\newblock A survey on image data augmentation for deep learning.
\newblock 6(1):60.

\bibitem{sanchez-gonzalez_median_2020}
José-María Sánchez-González, Carlos Rocha-de Lossada, and David Flikier.
\newblock Median absolute error and interquartile range as criteria of success
  against the percentage of eyes within a refractive target in {IOL} surgery.
\newblock 46(10):1441.

\bibitem{thirukumaran_missing_2012}
S.~Thirukumaran and A.~Sumathi.
\newblock Missing value imputation techniques depth survey and an imputation
  algorithm to improve the efficiency of imputation.
\newblock In {\em 2012 Fourth International Conference on Advanced Computing
  ({ICoAC})}, pages 1--5.
\newblock {ISSN}: 2377-6927.

\bibitem{tutz_improved_2015}
Gerhard Tutz and Shahla Ramzan.
\newblock Improved methods for the imputation of missing data by nearest
  neighbor methods.
\newblock 90:84--99.

\bibitem{veugen_privacy-preserving_2021}
Thijs Veugen, Bart Kamphorst, Natasja van~de L’Isle, and Marie~Beth van
  Egmond.
\newblock Privacy-preserving coupling of vertically-partitioned databases and
  subsequent training with gradient descent.
\newblock pages 38--51.

\bibitem{vinutha_detection_2018}
H.~P. Vinutha, B.~Poornima, and B.~M. Sagar.
\newblock Detection of outliers using interquartile range technique from
  intrusion dataset.
\newblock In Suresh~Chandra Satapathy, Joao Manuel~R.S. Tavares, Vikrant
  Bhateja, and J.~R. Mohanty, editors, {\em Information and Decision Sciences},
  Advances in Intelligent Systems and Computing, pages 511--518. Springer.

\bibitem{wan_estimating_2014}
Xiang Wan, Wenqian Wang, Jiming Liu, and Tiejun Tong.
\newblock Estimating the sample mean and standard deviation from the sample
  size, median, range and/or interquartile range.
\newblock 14(1):135.

\bibitem{wang_pc-gain_2021}
Yufeng Wang, Dan Li, Xiang Li, and Min Yang.
\newblock {PC}-{GAIN}: Pseudo-label conditional generative adversarial
  imputation networks for incomplete data.
\newblock 141:395--403.

\bibitem{yoon_gain_2018}
Jinsung Yoon, James Jordon, and Mihaela van~der Schaar.
\newblock {GAIN}: Missing data imputation using generative adversarial nets.

\bibitem{zhang_missing_2021}
Weibin Zhang, Pulin Zhang, Yinghao Yu, Xiying Li, Salvatore~Antonio Biancardo,
  and Junyi Zhang.
\newblock Missing data repairs for traffic flow with self-attention generative
  adversarial imputation net.
\newblock pages 1--12.
\newblock Conference Name: {IEEE} Transactions on Intelligent Transportation
  Systems.

\end{thebibliography}

\end{multicols}%
\end{document}